\documentclass[11pt]{article}
\usepackage[T1]{fontenc}
\usepackage[utf8]{inputenc}
\usepackage{authblk}

\usepackage{pdflscape}

\usepackage{amsmath, amssymb}
\usepackage{graphicx}
\begin{document}

	\title{Towards Identification of Relevant Variables in the observed Aerosol Optical Depth Bias between MODIS and AERONET observations}


\author[1]{N. K. Malakar }
\author[1]{ D. J. Lary}
\author[1]{D. Gencaga} 
\author[2]{\\A. Albayrak}
\author[2]{ J. Wei}
\affil[1]{Department of Physics, University of Texas at Dallas}
\affil[2]{Goddard Earth Sciences DISC, NASA Goddard Space Flight Center\\
 http://disc.gsfc.nasa.gov}
%


\maketitle
\begin{abstract}
Comparison of the aerosol optical depth values from two datasets, observed by satellite remote sensing Moderate Resolution Imaging Spectroradiometer (MODIS), and  globally distributed Aerosol Robotic Network (AERONET),  show that there are biases between the two data products.
In this paper, we present a general framework to identify the possible factors influencing the bias, which might be associated with the measurement conditions such as the solar and sensor zenith angles, the solar and sensor azimuth, scattering angles, and surface reflectivity at the various measured wavelengths, etc. Specifically, we performed analysis for remote sensing MODIS Aqua-Land data set, and used machine learning technique, neural network in this case, to perform multivariate  regression between the ground-truth and the training data sets.
Finally, we used mutual information between the observed and the predicted values as the measure of similarity. The set is then identified as the most relevant set of variables. The search consists of a brute force method as we have to consider all possible combinations of the regressors to the neural network. The computations involves a huge number crunching exercise, and we implemented it by writing a job-parallel program.

	\end{abstract}

\section*{Introduction}
 Atmospheric aerosols play an important role in earth's climate system \cite{pachauri2007ipcc}, and can also pose harmful effects on human health when inhaled. Therefore, accurately characterizing the global aerosol distribution is valuable for many reasons.

The total light extinction caused by aerosols over a vertical column in the atmosphere of unit cross section is  known as the aerosol optical depth (AOD) \cite{amt2012}.  Much effort has been placed in observing AOD from space and ground-based instruments \cite{Holben92, kaufman_operational_1997, Torres98,  holben_aeronet_1998, Mishchenko99, ichoku2002, remer_global_2008,  remer_modis_2005, chu2002validation, Liu2004}. 

\begin{figure}[tb]
\centering
\includegraphics[width=0.7\textwidth,angle=0]{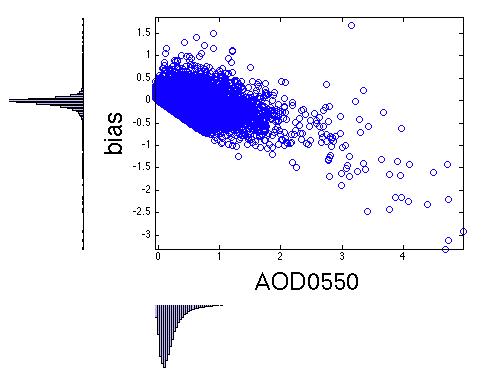}
   \caption{The distribution of bias with respect to the AOD at 550 nm. Although the bias at high AOD values are higher, the bias at the low AOD values are also important.}
   \label{fig:aodBias}
\end{figure}

 
A comparison between the AOD measurements made by MODIS and AERONET shows that there are biases between the two observations. Figure \ref{fig:aodBias} shows the distribution of the bias between the AOD measured by MODIS instrument at 550 nm. In this paper, we   will delineate our attempts to understand or explain the factors behind the bias.  We  performed a comprehensive brute force search for all possible combination of variables, and used neural network as one of the machine learning toolbox to predict the AOD values. Moreover, we used mutual information between the predicted and observed variables as the measure of correlation between them. We found the set which reproduced the AOD, producing  the highest value of mutual information with respect to the AERONET data set. The set is then identified as the most relevant set of variables for training the machine learning algorithm. The method of identification of relevant set of variable is a more general problem, which could be applied to similar multi-variate problems. Therefore, the presented example should be viewed as a test study. However, the framework and implementation will find its use for general purpose search for relevant variables.
 
 \section*{Data and Methodology}
 The data used in this study were derived from Multi-sensor Aerosol Products Sampling System (MAPSS), which derives the data from multiple sources such as original MODIS and AERONET datasets and provides level-2 aerosol scientific data sets \cite{amt2012, ichoku2002}. 
 MAPSS provides a consistent and uniform sampling of aerosol products by identifying the MODIS data pixels within approximately 27.5 km of the AERONET sites.   We analyzed the MODIS-aqua land data set. Readers interested in the details of the MAPSS  are directed to  \cite{amt2012}.   In the present paper we present the analysis of only the MODIS-Aqua land data set.  
 
 We applied a machine learning technique to predict the AOD values given the other components as the regressors to the learning module.  The method developed in this paper is constructed with plug-and-play in mind. The machine-learning module can be any regression tools of choice. In our case, we simply choose, as an example, Neural networks (NN) method. NNs are widely used in pattern recognition, machine learning and artificial intelligence.  In addition, NNs have found many applications in other fields such as geoscience, remote sensing, oceanography, etc. 
 
\begin{figure}[tb]
\centering
\includegraphics[width=0.5\textwidth,angle=0]{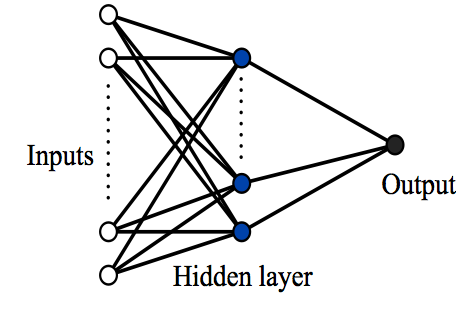}
   \caption{A cartoon of neural network, showing the inputs, hidden layers and output layers.  The machine learning module in this case consists of a neural network. }
   \label{fig:nnets}
\end{figure}

While training the NN as the regression tool, we regard  an observation data set as the product of n input variables, say $\{x_1, x_2, x_3, ... , x_n\}$, feeding into the regression machine. Therefore, the observed output variable, AOD, is some function of these input variables.  In terms of neural networks, as shown in figure \ref{fig:nnets}, the output of the $k^{th}$ neuron can be written as the weighted sum of inputs
\begin{equation}
y_k =  \varphi \left( \sum_{j=1}^n w_{kj} x_j \right),
\end{equation}
where  $\varphi$ is the transfer function, $w_{kj}$ represents the weight from unit $j$ to unit $k$, and $x_j$ represents the $n$ input variables to the neuron.
During training, the NN weights are adjusted appropriately to learn the data. The learning and adjustments of the weights are inspired by the synaptic learning behavior of neurons. 
The learning module is explored with all possible combinations of the input variables. 

The following variables have been used as the regressor to the neural network. We constructed all possible set of variables, this came out to be total of  32,781 possible combinations.
\begin{enumerate}
 \item	Aerosol optical depth at 550 nm (AOD0550) 
\item	Aerosol optical depth at 470 nm (AOD0470) 
\item	Aerosol optical depth at 660 nm (AOD0660) 
\item	Mean reflectance at 470 nm (mref0470) 
\item	Mean reflectance at 550 nm (mref0550) 
\item	Surface reflectance at 660 nm (surfre0660) 
\item	Surface reflectance at 470 nm (surfre0470) 
\item	Surface reflectance at 2100 nm (surfre2100) 
\item	Cloud fraction from land aerosol cloud mask (cfrac) 
\item	Quality assurance (QAavg) 
\item	Solar zenith angle (SolarZenith) 
\item	Solar azimuth angle (SolarAzimuth) 
\item	Viewing zenith angle (SensorZenith) 
\item	Sensor azimuth angle (SensorAzimuth) 
\item	Scattering angle (ScatteringAngle)
\end{enumerate}

For each combination set, one at a time, we trained the NN with AERONET AOD as the target variable. Once training phase is completed, we then predicted   the AERONET AOD  values from the trained network.  
The NN algorithm used a feed-forward back propagation algorithm with a hidden layer having 200 nodes.   The training was done by the Levenberg-Marquardt algorithm with mean-squared error as the performance factor. We used the Matlab toolbox.  We randomly split the training data set into three portions. The first $80\%$ portion is used to train the NN weights using an iterative process so that for each iteration, the  root mean square (RMS) error of the neural network is computed by using the second $10\%$ portion of the data. We used the RMS error to determine the convergence of our training. When the training is complete, we use the final $10\%$ of the data as the validation data set.    
 
Once the  neural network training is completed, we have obtained a mapping between the set of input and output variables. Therefore, the most relevant set of variables is the one that can best reproduce the target data.  We explored all combinations of variables, trained each of the combination, and compared which provided the fit of the observed AERONET AOD data. The end product is the result of regression between the input variables to the NN and the observed AERONET AOD. 
 
 The measure of similarity between predicted and observed AOD is measured by mutual information, which is described in next section. Since the learning module  constructs a mapping between the set of input and output variables, the most relevant set of variables is the one that can best reproduce the target data.

   \section*{Mutual Information}
The Correlation coefficient (Pearson's correlation) is a widely used measure of dependence between two variables, and represents the normalized measure of the strength of their linear relationships.   
The correlation coefficient $\rho_{X, Y}$ between two random variables $X$ and $Y$ with expected values $\mu_X$ and $\mu_Y$ and standard deviations $\sigma_X$  and $\sigma_Y$ is defined as
 \begin{eqnarray}
\rho_{X,Y}={\mathrm{cov}(X,Y) \over \sigma_X \sigma_Y} ={E[(X-\mu_X)(Y-\mu_Y)] \over \sigma_X\sigma_Y}
 \end{eqnarray}
%
where, E is the expected value operator, cov means covariance, and, $\rho$ a widely used alternative notation for Pearson's correlation.

The correlation coefficient is defined only if both of the standard deviations are finite and both of them are nonzero.  The correlation coefficients range from -1 to 1. The correlation coefficient values close to 1 (or -1) suggest that there is a positive (or negative) linear relationship between the data columns, whereas the values close to or equal to 0 suggest there is no linear relationship between the data columns. It can only be applied to the cases of linear relationship between two variables. 

Mutual information quantifies the mutual dependence between two variables by taking into account all of the characteristics of the variables in the Probability Distribution Function (PDF). Mutual information is defined as follows in discrete form:
\begin{equation}
   I(X,Y) = \sum_{x \in X} \sum_{y \in Y} p(x,y) \log \frac{p(x,y)}{p(x) p(y)},
   \end{equation}
 which is a special case of a measure called Kullback-Leibler divergence \cite{kullback_information_1997, cover_elements_2006}.
If $X$ and $Y$ are statistically independent, then 
\begin{equation}
p(x, y) =p(x) p(y).  
   \end{equation}
In this case, the mutual information becomes 0, showing independency. A proper mapping of the form
\begin{equation}
\delta(X,Y) = \sqrt{1- e^{-2I(X,Y)} }
   \end{equation}
normalizes the measure of general correlation as depicted by the MI \cite{joe_relative_1989, granger_using_1994, dionisio_mutual_2004}. 

In the case when X and Y are normally distributed, 
\begin{equation}
 (X,Y) ~ \sim \mathcal{N}(\mu, K)
    \end{equation}
where, 
$K = (\sigma^2, ~ \rho \sigma^2; \rho \sigma^2, ~ \sigma^2).$
Then, the mutual information reduces to 
\begin{equation}
I(X,Y)  = -\frac{1}{2} \log (1- \rho^2) .
   \end{equation}
So, that, 
 \begin{eqnarray}
  \delta(X,Y) = \sqrt{1- e^{-2I(X,Y)} } = |\rho(X,Y)|. \end{eqnarray}
This relation shows the generality of the normalized correlation measure.  
 
 There are several methods to estimate MI from data \cite{moon1995estimation, Reshef16122011, DenizME12}. We applied the Variable Bin Width Histogram Approach \cite{darbellay1999estimation, mutingCode}  to compute the  MI between the observed and predicted AOD.  Higher values indicate better agreement between the observed and predicted set, and thus are the best indicators of the input variables needed to assess a relevant set of variables.

 \section*{Results and Conclusions} 
We applied machine learning technique,  specifically neural network method in supervised learning mode, and explored all possible combination of variables. By the brute force exploration of all possible combinations, we found the best set of variables by comparing a measure of correlation between the predicted and observed variables. The measure of agreement between the observed data and the predicted data was obtained by using the mutual information. Therefore, we could identify the most relevant set of variables which could result into the maximum correlation between the predicted and observed data. 

\begin{figure}[tb]
\centering
\includegraphics[width=0.7\textwidth,angle=0]{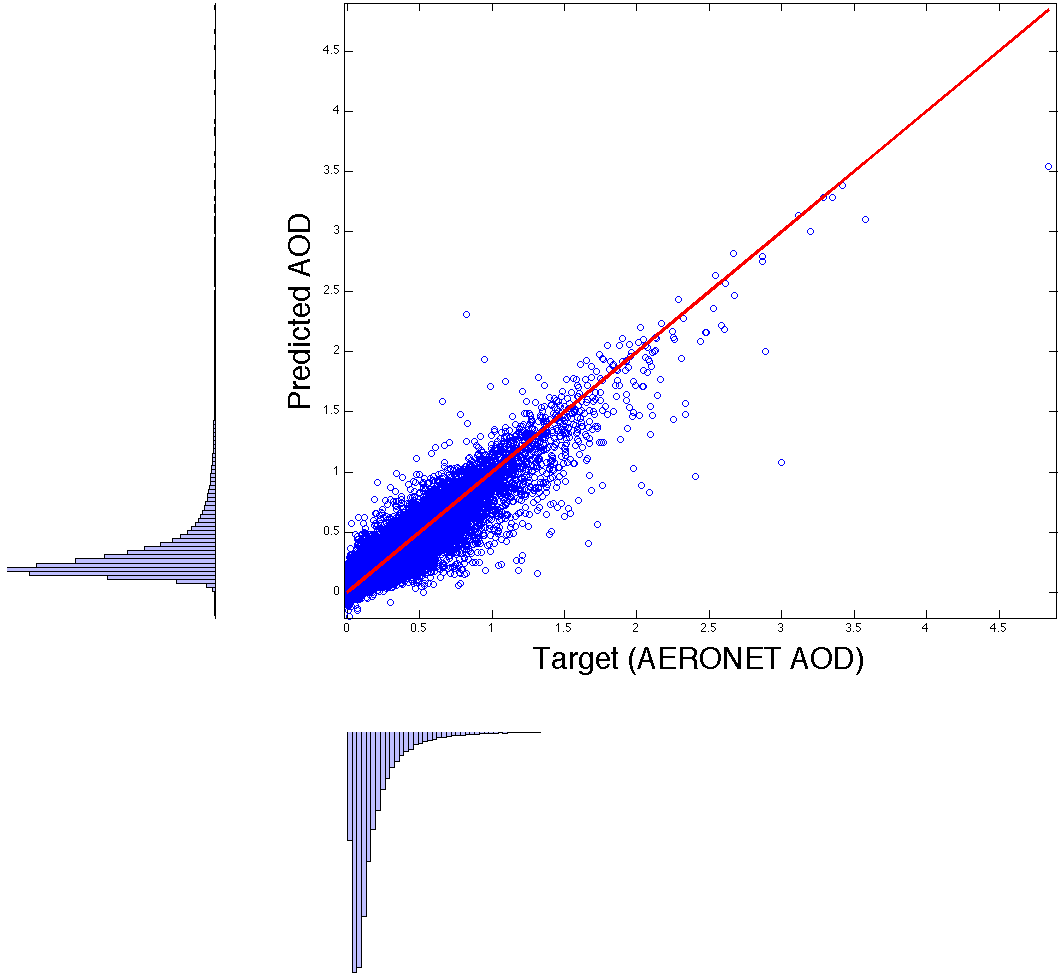}
   \caption{ The distribution of the AERONET AOD and predicted AOD after the NN training. The mutual information is $0.771$, and the correlation coefficient is $0.927$. This particular product was obtained by the best relevant set of variables as determined by the highest mutual information value between the two variables.}
   \label{fig:nnresults}
\end{figure}

Figure \ref{fig:nnresults} shows the best result we  obtained as a result of the brute force search.   It shows a comparison between the AERONET AOD, which was used as the target set for NN training, and the predicted AOD as the result of the NN training. The  mutual information and correlation coefficients values for the predicted and observed AOD values from the variable-set as regressor has been  presented in table \ref{table1}. For brevity we only show the first 15 rows of the table. The best prediction set had the MI value of $0.771$ to the AERONET AOD, and was result of  an input set  consisting of the following variables as regressors: AOD at 470 nm, and  AOD at 660 nm, mean reflectance at 470 nm, and mean reflectance at 550 nm, surface reflectance at 660nm, 470 nm, 2100 nm, cloud fraction, quality assurance values, solar zenith angle, solar azimuth angle, zenith angle, sensor azimuth angle and scattering angle.

There are several developments which could further benefit the methodology of finding relevant variables. In the future studies we will use mutual information with error bars.   In the future work we will present cross examination of multiple machine learning techniques to explain the bias-correction using the framework developed in this paper.

 We presented a brute force search method, which could be useful in many other cases involving multivariate exploration.  This could be useful in finding the most relevant set of factors to get insights from physical data. 

\begin{landscape}
\begin{table} 
\label{table1}
\caption{The top 15 results of brute force search is presented in the tabular format. The regressor variables are explained in the text.}
    \begin{tabular}{rrr}
        \hline
Combination & Mutual Information (MI)& Corr corrcoeff ($\rho$)\\ \hline    
$2,3,4,5,6,7,8,9,10,11,12,13,14,15$ &0.771&0.927\\
$1,2,4,5,6,7,8,10,11,12,13,15$&0.769&0.926\\
$1,2,3,4,5,6,8,10,11,12,13,14,15$&0.768&0.926\\
$1,2,4,5,6,8,9,10,11,12,13,14,15$&0.766&0.926\\
$1,2,3,4,5,6,7,8,9,10,12,13,15$&0.765&0.926\\
$1,2,4,5,7,8,10,12,13,14,15$&0.764&0.925\\
$1,2,4,5,6,7,8,9,10,11,12,13,14$&0.762&0.921\\
$2,3,4,5,6,7,8,10,11,12,13,14$&0.761&0.924\\
$1,3,4,5,6,7,8,10,11,12,13,14,15$&0.760&0.924\\
$1,2,4,5,7,8,10,11,12,13,14,15$&0.759&0.925\\
$1,2,4,5,6,7,10,11,12,13,14,15$&0.759&0.925\\
$1,3,4,5,6,7,8,9,10,11,12,13,15$&0.756&0.924\\
$1,2,4,5,6,8,10,11,12,13,15$&0.756&0.921\\
$1,2,4,5,7,8,10,11,12,13,15$&0.755&0.923\\
$2,3,4,5,7,8,9,10,11,12,13,15$&0.755&0.924\\
        \hline
    \end{tabular}
 \end{table}
\end{landscape}

\section*{Acknowledgements}
The support from NASA ACCESS AeroStat project (NNX10AM94G) is gratefully acknowledged. 
Special thanks are due to Dr. Maksym Petrenko and Dr. Charles Ichoku for providing Multi-sensor Aerosol Products Sampling System (MAPSS) data sets  \cite{amt2012}.
 We also thank the AERONET investigators, and the MODIS team for their wonderful work.

 \bibliographystyle{plain}
 \bibliography{sample}

\end{document}